\documentclass[letterpaper, 10 pt, conference]{ieeeconf}
\usepackage{color}

\IEEEoverridecommandlockouts

\usepackage{graphics} 
\usepackage{epsfig} 
\usepackage{amsmath} 
\usepackage{amssymb}  
\usepackage{algorithm}
\usepackage{algpseudocode}
\usepackage{lipsum}
\usepackage{xcolor}

\title{\LARGE \bf
    Leveraging Temporally Extended Behavior Sharing for Multi-task Reinforcement Learning
}

\author{
    Gawon Lee$^{1}$, Daesol Cho$^{2}$, H. Jin Kim$^{1\dagger}$
    \thanks{
        $^{1}$Department of Aerospace Engineering, Seoul National University.
    }
    \thanks{
        $^{2}$Artificial Intelligence Institute, Seoul National University.
    }
    \thanks{$^{\dagger}$Corresponding author.}
    \thanks{\tt\small \{lgw1997, dscho1234, hjinkim\}@snu.ac.kr}
}

\begin{document}

\maketitle
\thispagestyle{empty}
\pagestyle{empty}

\begin{abstract}
 
Multi-task reinforcement learning (MTRL) offers a promising approach to improve sample efficiency and generalization by training agents across multiple tasks, enabling knowledge sharing between them.
However, applying MTRL to robotics remains challenging due to the high cost of collecting diverse task data.
To address this, we propose MT-Lévy, a novel exploration strategy that enhances sample efficiency in MTRL environments by combining behavior sharing across tasks with temporally extended exploration inspired by Lévy flight \cite{levyflight}.
MT-Lévy leverages policies trained on related tasks to guide exploration towards key states, while dynamically adjusting exploration levels based on task success ratios.
This approach enables more efficient state-space coverage, even in complex robotics environments.
Empirical results demonstrate that MT-Lévy significantly improves exploration and sample efficiency, supported by quantitative and qualitative analyses.
Ablation studies further highlight the contribution of each component, showing that combining behavior sharing with adaptive exploration strategies can significantly improve the practicality of MTRL in robotics applications.
\end{abstract}

\section{INTRODUCTION}

Reinforcement learning (RL) aims to autonomously train an agent to solve complex control tasks in simulated and real-world environments \cite{IntroRL}.
Numerous successful applications of RL have been demonstrated in domains such as board games \cite{AlphaGo}, Atari games \cite{DQN}, and continuous control \cite{DDPG}.
However, applying RL to real-world tasks remains challenging because most RL algorithms require vast training data and are typically designed to solve only a single task.
To mitigate these issues of sample inefficiency and task specificity, multi-task RL (MTRL) has emerged as a promising research area \cite{MTRLsurvey, metaworld, MTRLQuadrotors}.
In MTRL, an agent is trained on multiple tasks simultaneously, enabling it to leverage knowledge acquired from one task to benefit learning in others.
This approach enhances sample efficiency and leads to more generalizable knowledge across tasks \cite{MOORE, CARE, PCGrad, QMP}.
Despite its benefits, implementing MTRL in robotics environments continues to encounter substantial hurdles.
Typical MTRL setups require a diverse set of tasks, often demanding intricate experimental configurations—such as deploying fleets of robots \cite{MT-opt} or employing automated reset mechanisms to collect data without human intervention \cite{ResetFree}.
While these approaches simplify data collection by scaling up experiments and reducing manual effort, they do not directly address the core issue: the inherent sample inefficiency of current MTRL algorithms.
From a robotics standpoint, reducing sample complexity is essential for making these algorithms practically viable.
One promising approach to improving sample efficiency is to adopt a more effective exploration mechanism \cite{RND, Count, ez-greedy}.
However, previous works have predominantly focused on exploration strategies for single-task environments, thus directly applying those strategies to MTRL settings can be inefficient and complex because they do not account for the presence of multiple tasks.
Thus, there is a growing need to develop an exploration method that is tailored to MTRL environments.
\begin{figure}
    \centering
    \includegraphics[width=1\linewidth]{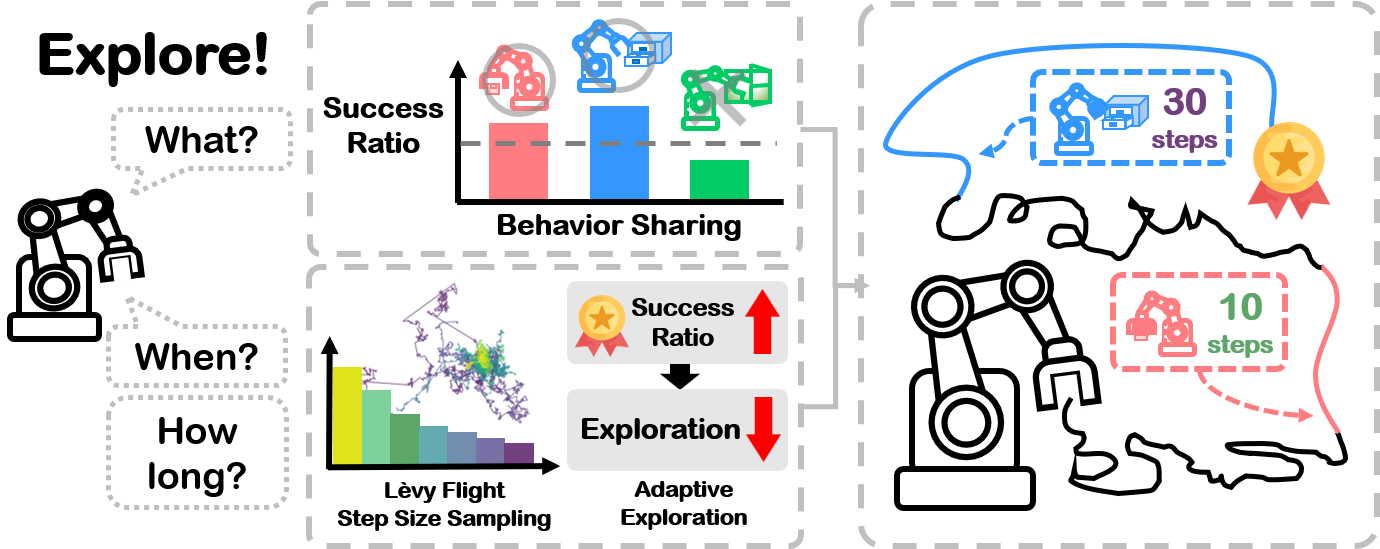}
    \caption{
        An overview of MT-Lévy. Our method improves exploration in MTRL by (1) sharing behaviors from related tasks, (2) executing exploratory actions for extended durations sampled from a Lévy flight distribution, and (3) adaptively tuning exploration parameters based on task success ratios.
    }
    \label{fig:mt-levy}
\end{figure}

To address the data collection challenge by improving sample efficiency, we propose three components that enhance exploration.
First, since tasks in MTRL commonly share features such as state/action spaces, behavior patterns, and reward structures, we propose leveraging policies trained on other tasks to facilitate exploration in a new task—a strategy known as behavior sharing \cite{QMP, CTPG}.
Although these policies might not yield optimal behavior for the new task, they tend to exhibit purposeful actions that effectively guide exploration \cite{MyopicMTRL}.
Second, executing these exploration policies in a temporally extended manner helps direct the agent toward states of interest \cite{ez-greedy, WhentoExp}.
In our approach, we employ Lévy flight \cite{levyflight}—a pattern observed in natural phenomena such as food foraging, transportation, and microbial movement—to implement this extended temporal exploration.
Finally, we utilize an exponential moving average of success ratios and track it to automatically adjust both the onset and duration of exploration.
This reduces the distribution mismatch between the data collection policy and the learned policy.
By integrating these components, we present MT-Lévy, which combines behavior sharing, temporally extended exploration using Lévy flight, and an automatic exploration adjustment mechanism based on success ratios.
Evaluations on MTRL benchmarks demonstrate that (1) MT-Lévy discovers more key states that are closely related to solving tasks, and (2) it achieves higher sample efficiency and performance than single-task oriented exploration methods and previously proposed MTRL algorithms.
In summary, our key contributions are as follows:
\begin{itemize}
    \item We propose MT-Lévy, an exploration method that enhances sample efficiency in MTRL environments by integrating behavior sharing, Lévy flight-based temporally extended exploration, and an automatic exploration adjustment mechanism.
    \item We empirically demonstrate that MT-Lévy facilitates more effective exploration by discovering key states, improves sample efficiency, and outperforms existing methods.
    Additionally, we provide comprehensive ablation studies to analyze the contribution of each component.
\end{itemize}
 
\section{RELATED WORKS}

\subsection{Multi-task Reinforcement Learning (MTRL)}

MTRL addresses two major barriers that limit the application of RL in real-world scenarios: high sample complexity and task specificity.
Unlike single-task RL algorithms, MTRL seeks to improve sample efficiency by sharing knowledge and achieving wider generalization, when learning from multiple tasks with similar characteristics.
\emph{Data Sharing}: This line of research involves sharing experience by relabeling data across tasks.
For example, \cite{IMPALA} uses parallelized training with multiple actors that share gradients with a central learner.
\cite{MT-opt} proposes to share data across tasks by relabeling and selecting the most appropriate transitions from the replay buffer, and \cite{CDS} builds on \cite{MT-opt} by incorporating conservative measures to control the effects of distributional shift arising from mixing the replay buffers across different tasks.
By sharing data, these methods aim to enhance the agent’s sample efficiency and generalization capabilities.
However, they do not explicitly address the exploration challenges that are crucial for further improving the sample efficiency of MTRL algorithms.
\emph{Parameter Sharing}: These methods share model parameters and adjust neural network architectures to facilitate knowledge transfer.
Approaches such as \cite{MOORE, CARE, PaCo, DynamicDepthRouting} share weights across tasks by modifying the neural network architecture or adopting a routing network.
However, this strategy can lead to negative interference when gradients from different tasks conflict.
Techniques such as \cite{PCGrad, CAGrad, NashMTL} have been proposed to alleviate such interference.
Works in this category are orthogonal to our approach, which is architecture-agnostic.
Thus, the proposed method could be used in tandem with these techniques to boost sample efficiency and performance.
\emph{Behavior Sharing}: Methods such as \cite{Actormimic, Distill} distill and regularize behaviors from agents trained on different tasks.
More recent works like \cite{QMP, CTPG} focus on explicitly sharing actions and are thus closely related to our work.
For instance, \cite{QMP} selects which action to use at each timestep by inferring the Q-function during exploration, which introduces an additional computational burden.
\cite{CTPG} introduces a high-level controller that selects actions from multiple task policies.
However, adopting a high-level controller can be numerically unstable due to off-policy corrections, and its control frequency is fixed, limiting adaptability.
In contrast, our work avoids the computational burden of inferring Q-values at each timestep or training an additional high-level controller, making it a simpler and more scalable exploration method tailored to MTRL environments.
\subsection{Exploration in RL}

Exploration is a fundamental challenge in RL, as agents must balance between exploiting known high-reward actions and exploring to discover novel, potentially rewarding states \cite{ez-greedy, RND}.
In many studies, naive methods such as $\epsilon$-greedy and Boltzmann exploration are adopted because of their simplicity and scalability \cite{AlphaGo, DQN, E2Evisuomotor}.
Although these methods often perform well enough, in hard-exploration problems—such as sparsely rewarded, high-dimensional environments—it can take exponentially many trials to reach a rewarding state \cite{MyopicMTRL}.
To overcome this exponential sample requirement, more sophisticated methods have been developed that encourage the agent to explore consistently rather than acting randomly at each timestep.
To promote more informed exploration, several approaches introduce intrinsic rewards based on prediction errors or state novelty \cite{Surprise, RND} or use information-theoretic measures to encourage information gain \cite{Information, MaxInfo}.
Additionally, temporally extended exploration methods—such as repeating the same randomly sampled action over multiple timesteps \cite{ez-greedy} or employing options that span several steps \cite{WhentoExp}—have shown promise.
However, these methods have predominantly been developed for single-task settings and do not consider the presence of multiple tasks when designing exploration strategies.
In our work, we adopt temporally extended exploration using Lévy flight and tailor the exploration strategy to better suit MTRL environments, leveraging the existence of multiple policies in multi-task scenarios.
\section{PRELIMINARIES}

\subsection{MTRL Formulation}

MTRL is defined over a set of Markov decision processes (MDPs) \cite{MDP}.
Specifically, let $\mathcal{M} = \{ \mathcal{M}_{1}, \mathcal{M}_{2}, \cdots, \mathcal{M}_{N} \}$ denote the set of MDPs, where an MDP is sampled from an arbitrary task distribution $\mathcal{T}$.
In most cases, $\mathcal{T}$ is a uniform distribution over the $N$ available tasks.
Each MDP in the set is a tuple $\langle \mathcal{S}, \mathcal{A}, T_{i}, R_{i} , \gamma, H \rangle$ where all MDPs share the state space $\mathcal{S}$ and action space $\mathcal{A}$ but differ in the state transition function $T: \mathcal{S} \times \mathcal{A} \to \mathcal{S}$ and the reward function $R: \mathcal{S} \times \mathcal{A} \to \mathbb{R}$.
Also, $\gamma$ and $H$ denote the discount factor and maximum trajectory length of the MDP, respectively.
MTRL aims to improve the sample efficiency and performance of an agent by sharing knowledge across tasks through learning a task-conditioned policy $\pi(a|s, i)=\pi_{i}$, where $i \in \{1, 2, \cdots, N\}$ is a task index representing one of $N$ different tasks.
Using the above notation, an MTRL problem can be expressed as the following optimization problem that finds the policy that maximizes the accumulated reward across all tasks in the MDP set:
\begin{equation}
    \pi^{*} 
    = \underset{\pi}{\text{argmax}}~ \mathbb{E}_{i \sim \mathcal{T}(\cdot)}
    \biggr[
        \mathbb{E}_{\pi_{i}} \Big[ \sum_{t=1}^{H} \gamma^{t-1} r_{t} \Big]
    \biggr]
\end{equation}
where $r_{t}$ is reward at time $t$.
Additionally, the behavior policy $\beta$ which collects transitions interacting with the environment often differs from the actual learned policy $\pi$.
We denote the true state distribution following each policy as $d_{\beta}$ and $d_{\pi}$, respectively.
\subsection{L{\'e}vy Flight}

\begin{figure}
    \centering
    \includegraphics[width=1\linewidth]{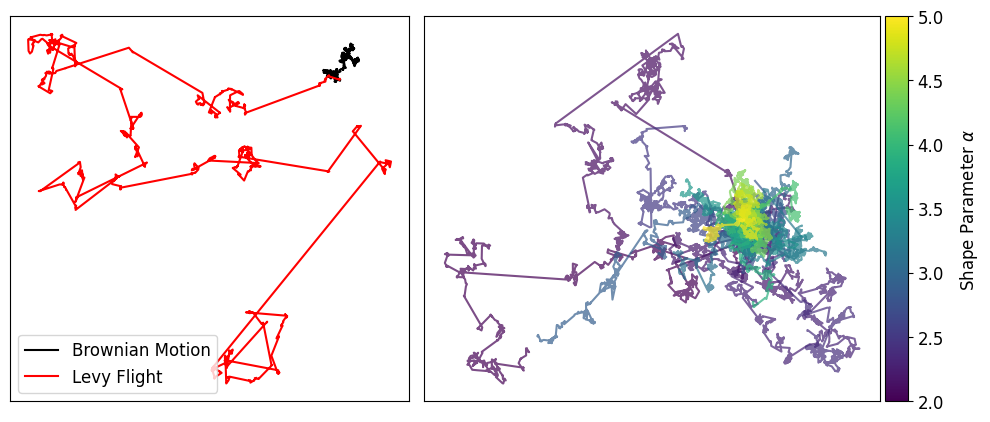}
    \caption{
        (\emph{left})
        Sampled 2D trajectories of L{\'e}vy flight and Brownian motion.
        Step sizes are sampled from Cauchy and normal distributions, respectively.
        Step directions are uniformly sampled from $[0, 2\pi)$.
        Lévy flight is characterized by sudden jumps in its trajectories, unlike Brownian motion.
        (\emph{right})
        Trajectories drawn by the step sizes sampled from Pareto distributions of different shape parameters $\alpha$.
        As $\alpha$ becomes smaller, the state coverage becomes wider.
    }
    \label{fig:levy-flight-traj}
\end{figure}

L{\'e}vy flight is a random walk in which step sizes are sampled from a stable, heavy-tailed probability distribution \cite{levyflight}.
It can be observed in a wide range of natural phenomena such as food foraging, transportation, and microbial movement \cite{WanderingAlbatrosses, Foodforaging, Bacteria}.
Figure \ref{fig:levy-flight-traj} (\emph{left}) illustrates sample trajectories of L{\'e}vy flight and Brownian motion.
In the case of L{\'e}vy flight, step sizes are drawn from a unit Cauchy distribution (a heavy-tailed distribution), whereas Brownian motion uses a unit Gaussian distribution for its step sizes.
The step directions are sampled uniformly from $[0, 2\pi)$. Consequently, the L{\'e}vy flight trajectory exhibits frequent local explorations interspersed with sudden long-range jumps, enabling it to cover a wider region compared to the trajectory of Brownian motion.
This property is particularly valuable in reinforcement learning, where the capability to cover large state spaces efficiently is crucial for improved learning performance \cite{ez-greedy, WhentoExp}.

Since we want to utilize Lévy flight to determine the exploration onset and duration, we adopt type-II Pareto distribution instead of the Cauchy distribution to ensure its support is positive.
This design choice is critical because exploration duration must be a positive value. Furthermore, the Pareto distribution not only has the desired heavy-tailed property characteristic of Lévy flights but also provides a shape parameter $\alpha$ that directly controls the heaviness of the tail. This allows for intuitive, one-dimensional control over the exploration-exploitation trade-off, where smaller $\alpha$ values lead to more aggressive exploration (longer, rarer "jumps") and larger values lead to more conservative behavior.
The type-II Pareto distribution is parameterized by a shape parameter $\alpha$ and a scale parameter $\lambda$.
Its probability density function is given by
\begin{equation} \label{pareto}
    \mathcal{P}(x; \alpha, \lambda) = \frac{\alpha}{\lambda}
    \bigl[ 1 + \frac{x}{\lambda} \bigr]^{-(\alpha + 1)}
    \quad (x > 0).
\end{equation}
Figure \ref{fig:levy-flight-traj} (\emph{right}) shows trajectories drawn using step sizes sampled from Pareto distributions with a fixed scale parameter $\lambda=1$ and varying shape parameters $\alpha$.
We can observe that by changing $\alpha$, we can adjust the strength of exploration, thereby adaptively balancing exploration and exploitation by modifying only a single parameter.
\section{METHODOLOGY}

In this section, we introduce three key components of our method: behavior sharing, temporally extended exploration using Lévy flight, and success rate tracking.
Our approach leverages behaviors acquired from other tasks to guide exploration and repeats exploratory actions for durations drawn from a Lévy flight, enabling the agent to efficiently traverse the state space.
Moreover, by continually monitoring the success rate, the algorithm automatically adjusts its exploration parameters, thereby balancing exploration and exploitation to improve sample efficiency.
\subsection{Behavior Sharing} \label{behavior_sharing}

\begin{figure}
    \centering
    \includegraphics[width=1\linewidth]{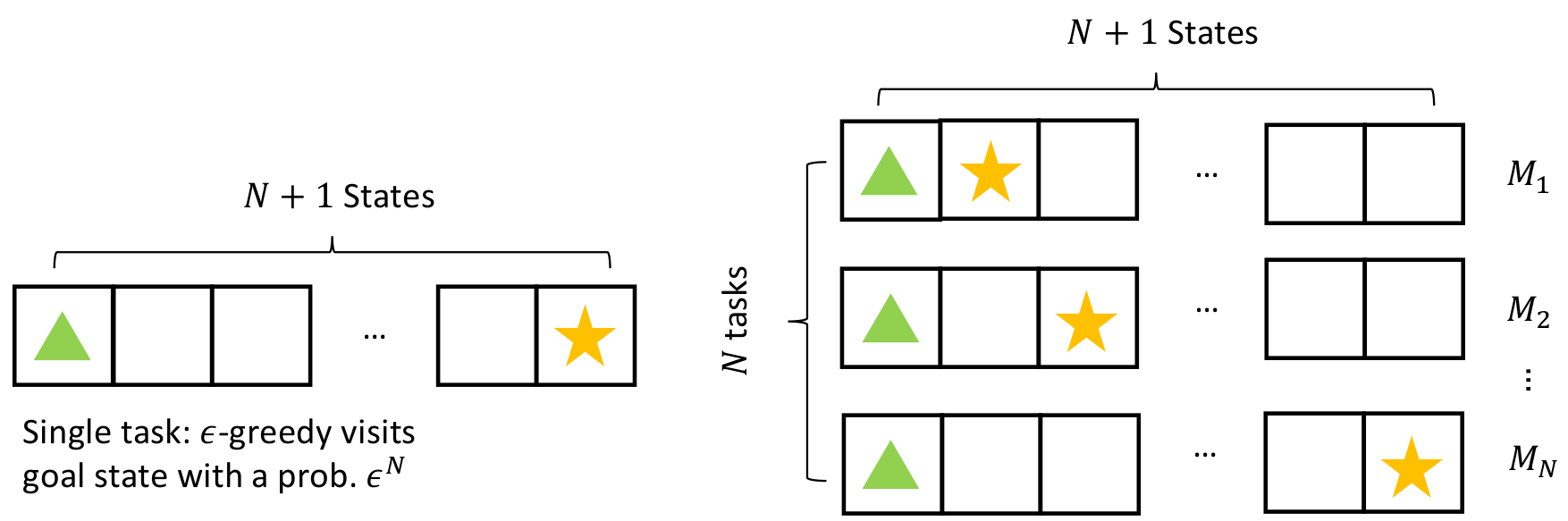}
    \caption{
        An agent is placed on a green triangle and the goal is to find a star placed on the right.
        The agent can only move in two directions: left or right.
        (\emph{left}) Single-task environment: the agent needs to traverse $N + 1$ states.
        (\emph{right}) Multi-task environment: the agent needs to traverse $i + 1$ states, depending on the task index $i$.
        Leveraging behavior sharing can greatly reduce the sample complexity in the multi-task environment.
        The figure is adapted from \cite{MyopicMTRL}.
    }
    \label{fig:behavior-sharing}
\end{figure}

To illustrate how behavior sharing can facilitate state-space exploration in multi-task environments, we present a didactic example in Figure \ref{fig:behavior-sharing}.
In the single-task case, employing a naive $\epsilon$-greedy exploration policy results in a sample complexity of $\mathcal{O}(1/\epsilon^{N})$.
Here, in the worst-case for this chain environment, the agent must take the correct exploratory action $N$ times consecutively to reach the goal, an event with probability proportional to $\epsilon^N$, leading to exponential sample complexity in the length of the chain.
By contrast, in the multi-task case, leveraging the policy learned in $\mathcal{M}_{i-1}$ for task $\mathcal{M}_{i}$ adds only an additional $\mathcal{O}(1/\epsilon)$ complexity, thereby reducing the overall sample requirement from exponential to linear scaling.

Based on this intuition, our method leverages policies conditioned on different task indices to roll out the current task under training.
Specifically, without behavior sharing, an action is sampled from the policy conditioned on the current task index $i$, $a_{i} \sim \pi_i(\cdot|s_i)$, where $s_i$ is the state collected in task $i$.
In contrast, when we explicitly share behavior across tasks, we condition the policy on another task index $j (\neq i)$, sampling an action $a_{j} \sim \pi_{j}(\cdot|s_i)$.
Although the state $s_i$ is likely to be unseen by the conditioned policy $\pi_{j}$, the shared actions can lead to key states that would be unlikely to be reached by following the current policy $\pi_{i}$ alone.
The task index $j (\neq i)$ is uniformly sampled from the set of tasks whose success ratios $\rho_i$ exceed the threshold $\bar{\rho}$.
However, this approach alone could sometimes select an index $j$ that hinders the exploration of the key states crucial for solving the task at hand.
Thus, we incorporate task metadata, such as task descriptions or task names, to guide behavior sharing toward tasks that exhibit sequentially similar behaviors, following \cite{CARE}.
By comparing the distances between task metadata embeddings and selecting only the closest ones, as described in Equation \ref{eqn:bs-set}, we choose indices that will effectively aid exploration.
\begin{equation} \label{eqn:bs-set}
    \mathcal{C}_{i, n}
    = \{ j \in \{ 1, \ldots, N\} ~|~ d(e_{i}, e_{j}) \leq d_{n} \}
\end{equation}
where $d(e_{i}, e_{j})=\|e_{i} - e_{j} \|_{2}$ is the L2 distance function and $d_{n}$ is the $n$-th smallest distance.
While using RoBERTa embeddings is effective, it introduces a dependency on a large pre-trained model; the embeddings, however, are computed offline once, imposing no overhead during training.

\subsection{Temporally Extended Exploration: L{\'e}vy Flight} \label{levy_flight}

Sampling actions from other policies alone does not sufficiently enhance exploration because a single action is unlikely to dramatically alter the trajectory.
Following \cite{ez-greedy}, we repeat the shared action for a prolonged duration so that it has a significant effect on the trajectory distribution.
Specifically, we first sample a timestep $c$ from a type-II Pareto distribution (Equation \ref{pareto})  with a fixed scale parameter $\lambda$ and an automatically adapted shape parameter $\alpha$ (as detailed in Section \ref{sr_tracking}).
The scale parameter $\lambda$ is fixed to 1, which simplifies the adaptation mechanism by allowing us to control the exploration strength exclusively through the shape parameter $\alpha$. This is deemed sufficient as $\alpha$ already provides direct and effective control over the distribution's heavy-tailedness across different tasks.
If $c > 1$, the agent enters exploration mode and samples an action from candidate exploration policies, repeating that action for $c$ timesteps.
If $c \leq 1$, the agent samples the action from the policy corresponding to the current task.
\subsection{Success Rate Tracking} \label{sr_tracking}

Although behavior sharing and temporally extended exploration using Lévy flight enhance sample efficiency, they can introduce a distribution mismatch that degrades performance \cite{CQL}.
In our method, the behavior policy $\beta_{i,\alpha}$ for task $i$ generates trajectories by stochastically initiating phases of temporally extended exploration, where actions from a shared policy are repeated, interspersed with actions from the task-specific policy $\pi_i$.
As a result, $\beta_{i, \alpha}$ can differ markedly from $\pi_{i}$, causing a severe mismatch between the empirical state distribution $\hat{d}_{\beta_{i, \alpha}}$ and the true state distribution $d_{\pi_{i}}$.
Moreover, naively fixing the distribution parameter $\alpha$ hinders the effective exploitation of learned experiences.
To address this issue, we track the success ratio for each task and use these ratios to regulate the distribution parameter $\alpha$, thereby reducing the mismatch and balancing exploration and exploitation.
The success ratio of a task is tracked as an exponentially moving average:
\begin{equation}
    \rho_{k+1}
    = (1 - \tau) \rho_{k} + \tau \varrho
\end{equation}
where $\rho_{k}$ is the success ratio after episode $k$, $\varrho$ is a binary flag set to 1 if the episode is successful (and 0 otherwise), and $\tau=0.01$ is the smoothing factor.
An episode is defined as ``successful'' based on the binary completion signal provided by the environment for each specific task.
The parameter $\alpha$ is then determined by
\begin{equation}
    \alpha
    = \bar{\alpha} + \frac{1}{\bar{\rho}^{(\rho/\bar{\rho})}}
\end{equation}
where $\bar{\alpha} = 1$ is a user-defined offset and $\bar{\rho}=0.1$ is the success ratio threshold.
As $\rho_{i}$ increases, $\alpha_{i}$ becomes larger, thereby encouraging the agent to exploit $\pi_{i}$.
Additionally, if the tracked success ratio exceeds $\bar{\rho}$, both behavior sharing and temporally extended exploration are disabled to prevent unnecessary exploration once the policy has nearly maximized the accumulated rewards.
The full overview of the proposed method is shown in Algorithm \ref{alg:mt-levy}.
The algorithm is run each environment step to decide whether to sample new action from the task policy or use actions that are previously sampled from other policies.

\begin{algorithm}
    \caption{MT-Lévy}
    \label{alg:mt-levy}
    \begin{algorithmic}[1]
        \Require $
            \{ \pi_{j} \}, \{ \rho_{j} \}, s, a, i, c, \mathcal{C}_{i, n}
        $
        \State $(c \gets 0)$
        \Comment{Only when an episode resets}
        
        \State $\alpha \gets \texttt{compute\_alpha}(\rho_{i}, \bar{\rho}, \bar{\alpha})$
        \Comment{\ref{sr_tracking}}
        
        \State $\mathcal{C} \gets \{ i \}$
        \Comment{Construct a set of candidate indices}
        \For{$j = 1, \ldots, N$}
            \If{$\rho_{j} > \bar{\rho} \text{ and } j \in \mathcal{C}_{i, n}$}
            \State $\mathcal{C} \cup \{ j \}$
            \Comment{\ref{behavior_sharing}}
            \EndIf
        \EndFor

        \If{$c \leq 1$}
            \State $c \sim \mathcal{P}(\cdot; \alpha, \lambda=1)$
            \Comment{\ref{levy_flight}}
            \If{$c > 1$}
                 \State $k \sim \texttt{uniform\_sample}(\mathcal{C})$
                 \State $a \sim \pi_{k}(\cdot|s)$
            \Else
                \State $a \sim \pi_{i}(\cdot|s)$
            \EndIf
        \EndIf
        \State $c \gets c - 1$
        \Comment{Decrement counter for repeated action}
        \State \Return $a$, $c$
    \end{algorithmic}
\end{algorithm}

\section{EXPERIMENTS}

To demonstrate that our method improves the sample efficiency and asymptotic performance of an MTRL algorithm, we conducted experiments in the multi-task manipulation environment MT10 \cite{metaworld}. In this environment, the agent simultaneously interacts with 10 different manipulation tasks (e.g., \texttt{reach}, \texttt{push}, and \texttt{pick-place}), all sharing the same state and action spaces.
In addition to the default dense reward functions provided with the tasks, we also conducted experiments using sparse reward functions that yield a reward of 1 when the agent completes a task.
The experiment in the sparse reward setting illustrates that our method enhances sample efficiency and performance even in hard-exploration problems.
Our method, MT-Lévy, is implemented on top of Multi-Head Soft Actor-Critic (MHSAC) \cite{metaworld}, a variant of Soft Actor-Critic (SAC) \cite{SAC} that attaches a separate head for each task.
For task metadata embeddings, we use 768-dimensional vectors obtained from the RoBERTa model \cite{Roberta}, as in \cite{CARE}.
Furthermore, we compare the proposed algorithm with Q-Mixture of Policies (QMP) \cite{QMP}, which shares behavior via a Q-switching mechanism to improve the efficiency of direct behavior sharing.
Additional hyperparameters used in the experiments are provided in the Appendix.
\subsection{Results}

\begin{figure}
    \centering
    \includegraphics[width=1\linewidth]{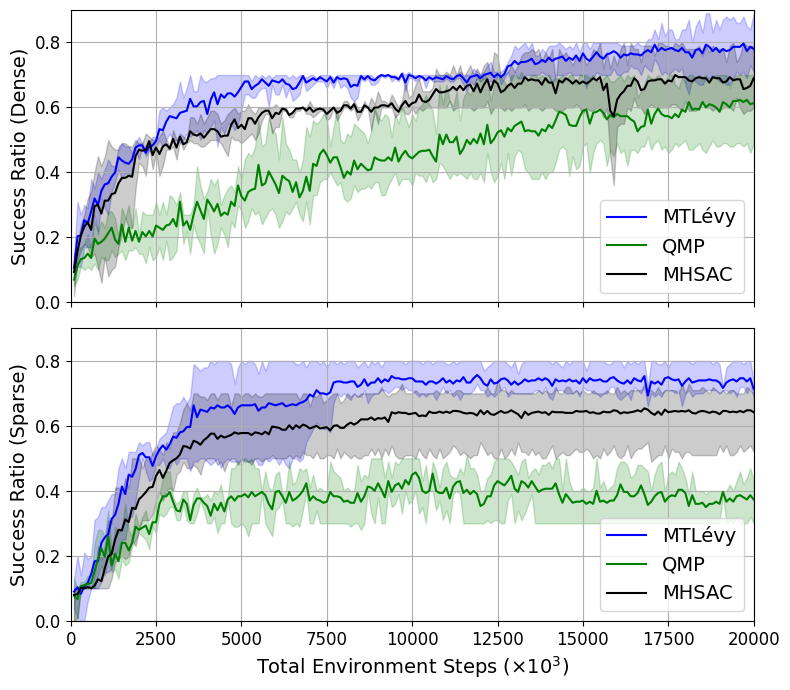}
    \caption{
        Training curves: MT10 (dense \& sparse).
        Trained for 20M total environment transitions. Means are aggregated over 3 different seeds, and colored ranges refer to maximum and minimum values.
        The proposed method, MT-Lévy, demonstrates superior sample efficiency and higher performance compared to baseline methods in both densely and sparsely rewarded environments.
    }
    \label{fig:main-experiment}
\end{figure}

The training curves for the agent in the MT10 dense and sparse environments are shown in Figure \ref{fig:main-experiment}.
In MHSAC, exploration relies on the standard SAC exploration, which is based on stochasticity from a Gaussian policy. This unstructured, per-timestep noise is often insufficient for meaningful exploration.
Furthermore, behaviors are shared indirectly through parameter sharing, resulting in reduced sample efficiency and performance due to the lack of exploration mechanisms that leverages explicit behavior sharing.
By contrast, QMP directly shares behaviors across tasks and thus more suited to exploration in the multi-task setting.
However, it determines which behavior to share based on Q-values, a strategy that can be detrimental during the early stages of training when the Q-functions are not informative.
The training result in the sparse environment highlights this issue, where meaningful information is scarce until the agent discovers rewards.
In contrast to these approaches, MT-Lévy promotes direct behavior sharing by using actions sampled from closely related tasks and repeating them, thereby enhancing exploration.
As a result, MT-Lévy achieves better sample efficiency and asymptotic performance, outperforming MHSAC by nearly 10 percentage points.

\begin{figure}
    \centering
    \includegraphics[width=1\linewidth]{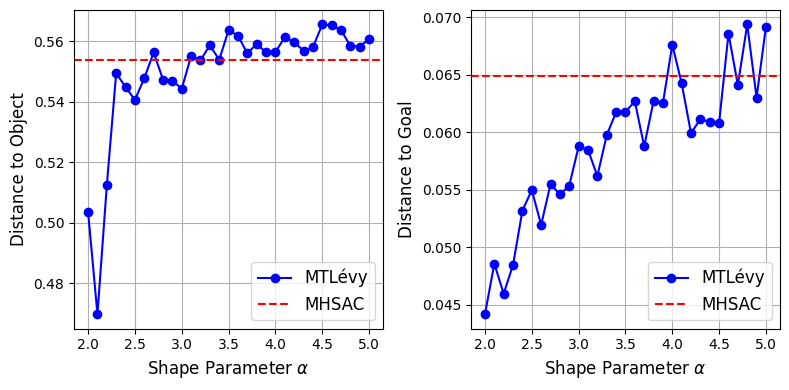}
    \caption{
        Quantitative analysis of key states in the \texttt{pick-place} environment.
        The plots show the mean of the lower 1\% of distances from the end-effector to the (\emph{left}) object position and (\emph{right}) goal position.
        As the shape parameter $\alpha$ (see Section \ref{levy_flight}) decreases, the agent explores more intensively, resulting in reduced mean distances to these key states.
    }
    \label{fig:object-gripper}
\end{figure}

In addition to the training curves, we present further quantitative and qualitative analyses of MT-Lévy.
To assess whether MT-Lévy promotes the exploration of key states crucial for task completion, we measured the mean of the lowest 1\% distances from the end-effector to both the object and the goal across all tasks in MT10.
We measured these distances throughout the training process before the success ratios of the tasks exceeded the threshold $\bar{\rho}$.
Figure \ref{fig:object-gripper} plots these mean distances, collected over 50 episodes per task at each training checkpoint, for varying values of $\alpha$ from 2 to 5. Compared to the original exploration strategy of MHSAC, MT-Lévy yields significantly smaller mean distances to both the object and the goal throughout training, indicating that our method encourages the discovery of key states essential for solving the tasks.

\begin{figure}
    \centering
    \includegraphics[width=1\linewidth]{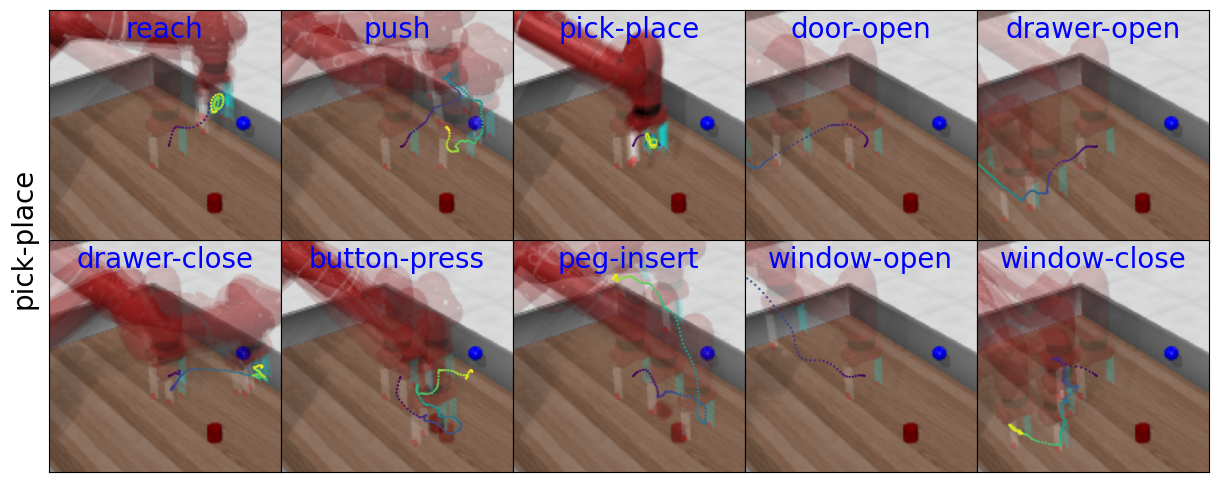}
    \includegraphics[width=1\linewidth]{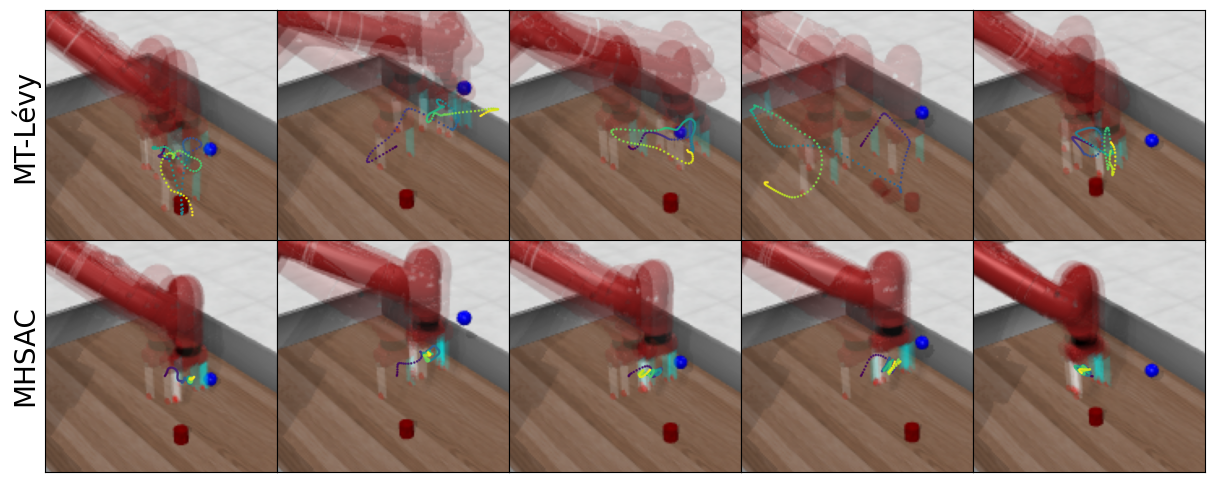}
    \caption{
        Qualitative analysis: 
        (\emph{top}) Example trajectories collected in the \texttt{pick-place} task by providing the policy with different task indices.
        The agent explores diverse key states even though some states are not observed during training.
        (\emph{bottom}) Example trajectories collected in the \texttt{pick-place} task using the proposed MT-Lévy method versus the baseline MHSAC exploration strategy.
        Our method encourages exploration through direct behavior sharing and temporally extended action execution.
        Color changes indicate the passage of timesteps, with purple marking the start and yellow marking the end.
    }
    \label{fig:exp-trajs}
\end{figure}

Furthermore, to qualitatively analyze the effect of MT-Lévy, we rendered example trajectories from the \texttt{pick-place} task.
In Figure \ref{fig:exp-trajs}, the upper 10 images depict trajectories obtained by applying alternative task indices to the policy.
Although these alternative policies do not yield directly interpretable trajectories, they nonetheless reveal key states in the task.
The lower panel compares trajectories generated using MT-Lévy with those produced by the base MHSAC exploration strategy.
These images illustrate that while the original exploration strategy of MHSAC often fails to explore the state space adequately, MT-Lévy encourages diverse behaviors such as interacting with the object or reaching for the goal, thereby enhancing the agent's exploration capability.

In terms of computational overhead, MT-Lévy adds negligible cost.
The main computational addition is the selection of an exploratory policy by sampling from a candidate indices set.
Although this introduces few more operations, it is more computationally efficient compared to the baseline QMP, as QMP requires computing a forward pass through the critic network once per action selection.
In addition, when in the exploration mode, the proposed algorithm does not incur any forward pass when selecting an action, as it simply repeats the previous action for a given duration.
Also, as mentioned in \ref{behavior_sharing}, calculation of embedding distances is performed only once offline and is negligible.

\subsection{Ablations}

\begin{figure}
    \centering
    \includegraphics[width=1\linewidth]{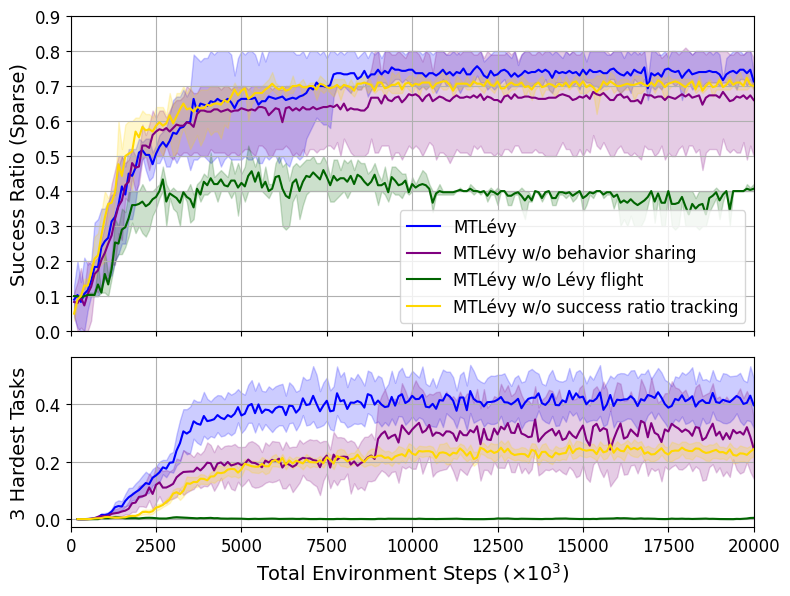}
    \caption{
        raining curves (ablation): MT10 (sparse).
        Trained for 20M total environment steps.
        Means are aggregated over 3 different seeds, and the colored ranges indicate the maximum and minimum values.
        (\emph{Top}) Mean success ratio across all 10 tasks, where the ablation variants consistently exhibit lower sample efficiency and performance compared to the proposed MT-Lévy method.
        (\emph{Bottom}) Mean success ratio for the three most challenging tasks: \texttt{push}, \texttt{pick-place}, and \texttt{door-open}.
        The performance gap becomes more pronounced after excluding the easier tasks.
    }
    \label{fig:ablation}
\end{figure}

To isolate the contribution of each component of MT-Lévy, we conducted ablation experiments in which we removed each component—behavior sharing, temporally extended exploration using Lévy flight, and success rate tracking—one at a time, and then trained the agent on the MT10 (sparse) environment.
In addition, we selected the three most challenging tasks from the set of 10 (namely, \texttt{push}, \texttt{pick-place}, and \texttt{door-open}) based on the observed success ratios during training, and plotted their training curves to more clearly demonstrate the effect of each component on the agent's exploration capability.
Although the \texttt{peg-insert-side} task was also challenging, we excluded it from our analysis because none of the ablation variants were able to solve it.
Removing the easier tasks from the analysis makes the impact of ablating each component even more pronounced.
First, removing behavior sharing results in an algorithm highly similar to $\epsilon$-greedy \cite{ez-greedy}, differing only in the distribution used to sample exploration step sizes.
These results demonstrate that incorporating behavior sharing on top of temporally extended exploration significantly improves sample efficiency.
Next, when temporally extended exploration using Lévy flight is omitted, the algorithm reverts to sharing behavior on a per-step basis, which yields the worst performance since it is nearly equivalent to random action sampling.
Finally, without success rate tracking, the agent cannot adaptively tune the onset and duration of exploration;
the results show that success ratio tracking provides additional performance gains by enabling the agent to exploit better the knowledge it has acquired.
Overall, our ablation study confirms that omitting even a single component of the proposed method leads to reduced sample efficiency and performance, underscoring the critical role of each component.
\section{CONCLUSIONS}

In this work, we introduced MT-Lévy, a novel exploration method tailored for MTRL in robotics.
By integrating behavior sharing, temporally extended exploration via Lévy flight, and success rate tracking that adapts the exploration parameter, our approach effectively guides the agent toward key states critical for task success.
Our experiments on the MT10 manipulation benchmark—under both dense and sparse reward conditions—demonstrate that MT-Lévy significantly improves sample efficiency and overall performance compared to existing methods.
Moreover, our ablation studies confirm that each component of MT-Lévy is essential to its effectiveness.

Currently, the mechanisms for selecting which behavior to share and for adapting hyperparameters rely on success ratio tracking, which is impractical when a clear criterion for task success is unavailable.
Consequently, future research could refine these components by incorporating, for example, large language models to provide more principled alternatives for behavior sharing and adaptive parameter tuning.
Such improvements have the potential to further enhance the robustness and generalizability of MTRL algorithms, ultimately paving the way for scalable and efficient robotic systems capable of operating in diverse real-world environments.



\section*{APPENDIX} \label{appendix}

Hyperparameter settings for MHSAC are detailed in Table \ref{tab:hyperparams}, while those specific to MT-Lévy are presented in Table \ref{tab:mtlevy-hyperparams}.

\begin{table}[ht]
    \centering
    \caption{MHSAC Hyperparameters}
    \label{tab:hyperparams}
    \begin{tabular}{lc}
        \hline
        \textbf{Parameter} & \textbf{Value} \\
        \hline
        Discount $\gamma$ & 0.99 \\
        \# heads & \# tasks $N$ \\
        \# hidden layers & 3 \\
        \# hidden units & 400 \\
        Optimizer & Adam \\
        Actor \& critic learning rate & 0.0003 \\
        Temperature learning rate & 0.0001 \\
        Actor \texttt{log\_std} bound & $[-20, 2]$ \\
        Critic target update rate & 0.005 \\
        Bound critic & False (dense), True (sparse) \\
        Initial temperature & 0.1 (dense), 0.001 (sparse) \\
        Replay buffer capacity & $10^{6}$ \\
        Batch size & $128 \times N$ \\
        \hline
    \end{tabular}
\end{table}

\begin{table}[ht]
    \centering
    \caption{MT-Lévy Hyperparemters}
    \label{tab:mtlevy-hyperparams}
    \begin{tabular}{lc}
        \hline
        \textbf{Parameter} & \textbf{Value} \\
        \hline
        $\alpha$ offset $\bar{\alpha}$ & 1 \\
        Success ratio threshold $\bar{\rho}$ & 0.1 \\
        Success ratio update rate $\tau$ & 0.01 \\
        \# closest neighbors $n$ & 5 \\
        \hline
    \end{tabular}
\end{table}

\section*{ACKNOWLEDGMENT}

This work was supported by Samsung Research Funding \& Incubation Center of Samsung Electronics under Project Number SRFC-IT2402-17.

\bibliographystyle{ieeetr}
\bibliography{IROS25}

\begin{thebibliography}{10}

\bibitem{levyflight}
B.~B. Mandelbrot, ``The fractal geometry of nature/revised and enlarged edition,'' {\em New York}, 1983.

\bibitem{IntroRL}
R.~S. Sutton, ``Reinforcement learning: An introduction,'' {\em A Bradford Book}, 2018.

\bibitem{AlphaGo}
D.~Silver, J.~Schrittwieser, K.~Simonyan, I.~Antonoglou, A.~Huang, A.~Guez, T.~Hubert, L.~Baker, M.~Lai, A.~Bolton, {\em et~al.}, ``Mastering the game of go without human knowledge,'' {\em nature}, vol.~550, no.~7676, pp.~354--359, 2017.

\bibitem{DQN}
V.~Mnih, ``Playing atari with deep reinforcement learning,'' {\em arXiv preprint arXiv:1312.5602}, 2013.

\bibitem{DDPG}
T.~Lillicrap, ``Continuous control with deep reinforcement learning,'' {\em arXiv preprint arXiv:1509.02971}, 2015.

\bibitem{MTRLsurvey}
N.~Vithayathil~Varghese and Q.~H. Mahmoud, ``A survey of multi-task deep reinforcement learning,'' {\em Electronics}, vol.~9, no.~9, p.~1363, 2020.

\bibitem{metaworld}
T.~Yu, D.~Quillen, Z.~He, R.~Julian, K.~Hausman, C.~Finn, and S.~Levine, ``Meta-world: A benchmark and evaluation for multi-task and meta reinforcement learning,'' in {\em Conference on robot learning}, pp.~1094--1100, PMLR, 2020.

\bibitem{MTRLQuadrotors}
J.~Xing, I.~Geles, Y.~Song, E.~Aljalbout, and D.~Scaramuzza, ``Multi-task reinforcement learning for quadrotors,'' {\em IEEE Robotics and Automation Letters}, 2024.

\bibitem{MOORE}
A.~Hendawy, J.~Peters, and C.~D'Eramo, ``Multi-task reinforcement learning with mixture of orthogonal experts,'' {\em arXiv preprint arXiv:2311.11385}, 2023.

\bibitem{CARE}
S.~Sodhani, A.~Zhang, and J.~Pineau, ``Multi-task reinforcement learning with context-based representations,'' in {\em International Conference on Machine Learning}, pp.~9767--9779, PMLR, 2021.

\bibitem{PCGrad}
T.~Yu, S.~Kumar, A.~Gupta, S.~Levine, K.~Hausman, and C.~Finn, ``Gradient surgery for multi-task learning,'' {\em Advances in Neural Information Processing Systems}, vol.~33, pp.~5824--5836, 2020.

\bibitem{QMP}
G.~Zhang, A.~Jain, I.~Hwang, S.-H. Sun, and J.~J. Lim, ``{QMP}: Q-switch mixture of policies for multi-task behavior sharing,'' in {\em The Thirteenth International Conference on Learning Representations}, 2025.

\bibitem{MT-opt}
D.~Kalashnikov, J.~Varley, Y.~Chebotar, B.~Swanson, R.~Jonschkowski, C.~Finn, S.~Levine, and K.~Hausman, ``Mt-opt: Continuous multi-task robotic reinforcement learning at scale,'' {\em arXiv preprint arXiv:2104.08212}, 2021.

\bibitem{ResetFree}
A.~Gupta, J.~Yu, T.~Z. Zhao, V.~Kumar, A.~Rovinsky, K.~Xu, T.~Devlin, and S.~Levine, ``Reset-free reinforcement learning via multi-task learning: Learning dexterous manipulation behaviors without human intervention,'' in {\em 2021 IEEE International Conference on Robotics and Automation (ICRA)}, pp.~6664--6671, IEEE, 2021.

\bibitem{RND}
Y.~Burda, H.~Edwards, A.~Storkey, and O.~Klimov, ``Exploration by random network distillation,'' {\em arXiv preprint arXiv:1810.12894}, 2018.

\bibitem{Count}
H.~Tang, R.~Houthooft, D.~Foote, A.~Stooke, O.~Xi~Chen, Y.~Duan, J.~Schulman, F.~DeTurck, and P.~Abbeel, ``\# exploration: A study of count-based exploration for deep reinforcement learning,'' {\em Advances in neural information processing systems}, vol.~30, 2017.

\bibitem{ez-greedy}
W.~Dabney, G.~Ostrovski, and A.~Barreto, ``Temporally-extended $\{$$\backslash$epsilon$\}$-greedy exploration,'' {\em arXiv preprint arXiv:2006.01782}, 2020.

\bibitem{CTPG}
J.~He, K.~Li, Y.~Zang, H.~Fu, Q.~Fu, J.~Xing, and J.~Cheng, ``Efficient multi-task reinforcement learning with cross-task policy guidance,'' in {\em The Thirty-eighth Annual Conference on Neural Information Processing Systems}, 2024.

\bibitem{MyopicMTRL}
Z.~Xu, Z.~Xu, R.~Jiang, P.~Stone, and A.~Tewari, ``Sample efficient myopic exploration through multitask reinforcement learning with diverse tasks,'' {\em arXiv preprint arXiv:2403.01636}, 2024.

\bibitem{WhentoExp}
M.~Pislar, D.~Szepesvari, G.~Ostrovski, D.~Borsa, and T.~Schaul, ``When should agents explore?,'' {\em arXiv preprint arXiv:2108.11811}, 2021.

\bibitem{IMPALA}
L.~Espeholt, H.~Soyer, R.~Munos, K.~Simonyan, V.~Mnih, T.~Ward, Y.~Doron, V.~Firoiu, T.~Harley, I.~Dunning, {\em et~al.}, ``Impala: Scalable distributed deep-rl with importance weighted actor-learner architectures,'' in {\em International conference on machine learning}, pp.~1407--1416, PMLR, 2018.

\bibitem{CDS}
T.~Yu, A.~Kumar, Y.~Chebotar, K.~Hausman, S.~Levine, and C.~Finn, ``Conservative data sharing for multi-task offline reinforcement learning,'' {\em Advances in Neural Information Processing Systems}, vol.~34, pp.~11501--11516, 2021.

\bibitem{PaCo}
L.~Sun, H.~Zhang, W.~Xu, and M.~Tomizuka, ``Paco: Parameter-compositional multi-task reinforcement learning,'' {\em Advances in Neural Information Processing Systems}, vol.~35, pp.~21495--21507, 2022.

\bibitem{DynamicDepthRouting}
J.~He, K.~Li, Y.~Zang, H.~Fu, Q.~Fu, J.~Xing, and J.~Cheng, ``Not all tasks are equally difficult: Multi-task deep reinforcement learning with dynamic depth routing,'' in {\em Proceedings of the AAAI Conference on Artificial Intelligence}, vol.~38, pp.~12376--12384, 2024.

\bibitem{CAGrad}
B.~Liu, X.~Liu, X.~Jin, P.~Stone, and Q.~Liu, ``Conflict-averse gradient descent for multi-task learning,'' {\em Advances in Neural Information Processing Systems}, vol.~34, pp.~18878--18890, 2021.

\bibitem{NashMTL}
A.~Navon, A.~Shamsian, I.~Achituve, H.~Maron, K.~Kawaguchi, G.~Chechik, and E.~Fetaya, ``Multi-task learning as a bargaining game,'' {\em arXiv preprint arXiv:2202.01017}, 2022.

\bibitem{Actormimic}
E.~Parisotto, J.~L. Ba, and R.~Salakhutdinov, ``Actor-mimic: Deep multitask and transfer reinforcement learning,'' {\em arXiv preprint arXiv:1511.06342}, 2015.

\bibitem{Distill}
A.~A. Rusu, S.~G. Colmenarejo, C.~Gulcehre, G.~Desjardins, J.~Kirkpatrick, R.~Pascanu, V.~Mnih, K.~Kavukcuoglu, and R.~Hadsell, ``Policy distillation,'' {\em arXiv preprint arXiv:1511.06295}, 2015.

\bibitem{E2Evisuomotor}
S.~Levine, C.~Finn, T.~Darrell, and P.~Abbeel, ``End-to-end training of deep visuomotor policies,'' {\em Journal of Machine Learning Research}, vol.~17, no.~39, pp.~1--40, 2016.

\bibitem{Surprise}
J.~Achiam and S.~Sastry, ``Surprise-based intrinsic motivation for deep reinforcement learning,'' {\em arXiv preprint arXiv:1703.01732}, 2017.

\bibitem{Information}
S.~Still and D.~Precup, ``An information-theoretic approach to curiosity-driven reinforcement learning,'' {\em Theory in Biosciences}, vol.~131, pp.~139--148, 2012.

\bibitem{MaxInfo}
B.~Sukhija, S.~Coros, A.~Krause, P.~Abbeel, and C.~Sferrazza, ``Maxinforl: Boosting exploration in reinforcement learning through information gain maximization,'' {\em arXiv preprint arXiv:2412.12098}, 2024.

\bibitem{MDP}
M.~L. Puterman, {\em Markov decision processes: discrete stochastic dynamic programming}.
\newblock John Wiley \& Sons, 2014.

\bibitem{WanderingAlbatrosses}
G.~M. Viswanathan, V.~Afanasyev, S.~V. Buldyrev, E.~J. Murphy, P.~A. Prince, and H.~E. Stanley, ``L{\'e}vy flight search patterns of wandering albatrosses,'' {\em Nature}, vol.~381, no.~6581, pp.~413--415, 1996.

\bibitem{Foodforaging}
D.~A. Raichlen, B.~M. Wood, A.~D. Gordon, A.~Z. Mabulla, F.~W. Marlowe, and H.~Pontzer, ``Evidence of l{\'e}vy walk foraging patterns in human hunter--gatherers,'' {\em Proceedings of the National Academy of Sciences}, vol.~111, no.~2, pp.~728--733, 2014.

\bibitem{Bacteria}
A.~Be’er and G.~Ariel, ``A statistical physics view of swarming bacteria,'' {\em Movement ecology}, vol.~7, pp.~1--17, 2019.

\bibitem{CQL}
A.~Kumar, A.~Zhou, G.~Tucker, and S.~Levine, ``Conservative q-learning for offline reinforcement learning,'' {\em Advances in neural information processing systems}, vol.~33, pp.~1179--1191, 2020.

\bibitem{SAC}
T.~Haarnoja, A.~Zhou, K.~Hartikainen, G.~Tucker, S.~Ha, J.~Tan, V.~Kumar, H.~Zhu, A.~Gupta, P.~Abbeel, {\em et~al.}, ``Soft actor-critic algorithms and applications,'' {\em arXiv preprint arXiv:1812.05905}, 2018.

\bibitem{Roberta}
Y.~Liu, M.~Ott, N.~Goyal, J.~Du, M.~Joshi, D.~Chen, O.~Levy, M.~Lewis, L.~Zettlemoyer, and V.~Stoyanov, ``Roberta: A robustly optimized bert pretraining approach,'' {\em arXiv preprint arXiv:1907.11692}, 2019.

\end{thebibliography}

\end{document}